\documentclass{llncs} 
\usepackage{url,alltt}
\usepackage{amssymb}

\usepackage[arrow, matrix, curve]{xy} 

\usepackage{epsfig}
\usepackage{wrapfig}
\usepackage{subfigure}
\usepackage{rotating}
\usepackage{moreverb}
\usepackage{fancyvrb}

\def\bd{\begin{description}}
\def\ed{\end{description}}
\def\bc{\begin{center}}
\def\ec{\end{center}}
\def\bq{\begin{quote}}
\def\eq{\end{quote}}
\def\bi{\begin{itemize}}
\def\ei{\end{itemize}}
\def\be{\begin{enumerate}}
\def\ee{\end{enumerate}}
\def\ba{\begin{array}}
\def\ea{\end{array}}

\newcommand{\chr}{{CHR}}
\newcommand{\true}{{\it true}}
\newcommand{\equal}{{\tt =}}
\newcommand{\false}{{\it false}}

\newcommand{\myparagraph}[1]{\textbf{#1.}}

\begin{document}

\title{Justifications in Constraint Handling Rules for Logical Retraction in Dynamic Algorithms} 
\author{%
Thom Fr{\"u}hwirth
}
\institute{%
Ulm University, Germany\\
\email{thom.fruehwirth@uni-ulm.de}
}

\maketitle

\begin{abstract}
We present a straightforward source-to-source transformation that introduces 
justifications for user-defined constraints into the CHR programming language.
Then a scheme of two rules suffices to allow for logical retraction (deletion, removal) of constraints during computation.
Without the need to recompute from scratch, these rules remove not only the constraint but also undo all consequences of the rule applications that involved the constraint. 
We prove a confluence result concerning the rule scheme and show its correctness.

When algorithms are written in CHR, constraints represent both data and operations. 
CHR is already incremental by nature, i.e. constraints can be added at runtime.
Logical retraction adds decrementality.
Hence any algorithm written in CHR with justifications will become fully dynamic. 
Operations can be undone and data can be removed at any point in the computation without compromising the correctness of the result.

We present two classical examples of dynamic algorithms,
written in our prototype implementation of CHR with justifications that is available online:
maintaining the minimum of a changing set of numbers and shortest paths in a graph whose edges change.
\end{abstract}

\section{Introduction}

Justifications have their origin in truth maintenance systems (TMS) \cite{mcallester1990truth} for automated reasoning.
In this knowledge representation method, derived information (a formula) is explicitly stored and associated with the information it originates from by means of justifications. 
This dependency can be used to explain the reason for a conclusion (consequence) by its initial premises. 
With the help of justifications, conclusions can be withdrawn by retracting their premises. 
By this {\em logical retraction}, e.g. default reasoning can be supported and 
inconsistencies can be repaired by retracting one of the reasons for the inconsistency.
An obvious application of justifications are dynamic constraint satisfaction problems (DCSP), in particular over-constrained ones \cite{brown2006uncertainty}.

In this work, we extend the applicability of logical retraction to arbitrary algorithms
that are expressed in the programming language Constraint Handling Rules (CHR) \cite{chrbook,fruhwirth2015constraint}.
To accomplish logical retraction, we have to be aware that
CHR constraints can also be deleted by rule applications.
These constraints may have to be restored when a premise is retracted.
With logical retraction, any algorithm written in CHR will become {\em fully dynamic\footnote{Dynamic algorithms for dynamic problems should not be confused with dynamic programming.}}.

\myparagraph{Minimum Example}
Given a multiset of numbers {\tt min($n_1$), min($n_2$),\ldots, min($n_k$)}.
The constraint (predicate) {\tt min($n_i$)} means that the number $n_i$ is a candidate
for the minimum value.
The following CHR rule filters the candidates.
\begin{verbatim}
min(N) \ min(M) <=> N=<M | true.
\end{verbatim}
The rule consists of a left-hand side, on which a pair of constraints has to be matched,
a guard check {\tt N=<M} that has to be satisfied, and an empty right-hand side denoted by {\tt true}.
In effect, the rule takes two {\tt min} candidates and removes the one with the
larger value (constraints after the \verb+\+ symbol are deleted). 
Note that the {\tt min} constraints behave both as operations (removing other constraints) and as data (being removed).

CHR rules are applied exhaustively.
Here the rule keeps on going until only one, thus the smallest
value, remains as single {\tt min} constraint, denoting the current minimum.
If another {\tt min} constraint is added during the computation, it will eventually react with a previous {\em min} constraint, and the correct current minimum will be computed in the end. Thus the algorithm as implemented in CHR is incremental.
It is not decremental, though: We cannot logically retract a {\em min} candidate. 
While removing a candidate that is larger than the minimum would be trivial, 
the retraction of the minimum itself requires to remember all deleted candidates and to find their minimum.
With the help of justifications, this logical retraction will be possible automatically.

{\bf Contributions and Overview of the Paper.}
In the next section we recall syntax and operational semantics for CHR.
Our contributions are as follows:
\begin{itemize}

\item We introduce CHR with justifications (CHR$^{\mathcal J}$) in Section 3. 
We enhance standard CHR programs with justifications by a source-to-source program transformation. 
We show the operational equivalence of rule applications in both settings.
Thus CHR$^{\mathcal J}$ is a conservative extension of standard CHR.

\item We define a scheme of two rules to enable logical retraction of constraints based on justifications in Section 4.
We show that the rule scheme is confluent with each rule in any given program, independent of the confluence of that program.
We prove correctness of logical retraction:
the result of a computation with retraction is the same as if the constraint would never have been introduced in the computation.

\item We present a proof-of-concept implementation of CHR$^{\mathcal J}$ in CHR and Prolog (available online) in Section 5.
We discuss two classical examples for dynamic algorithms,
maintaining the minimum of a changing set of numbers and 
maintaining shortest paths in a graph whose edges change.
\end{itemize}
The paper ends with discussion of related work in Section 6 and with conclusions and directions for future work.

\section{Preliminaries}\label{sec:chr}

We recall the abstract syntax and the equivalence-based abstract operational semantics of CHR in this section.
Upper-case letters stand for (possibly empty) conjunctions of constraints in this paper.

\subsection{Abstract Syntax of CHR}

{\em Constraints} are relations, distinguished predicates of first-order predicate logic.
We differentiate between 
two kinds of constraints: {\em built-in (pre-defined) constraints} 
and
{\em user-defined (CHR) constraints} 
which are defined by the rules in a CHR program.
\begin{definition}
{\rm
A {\em \chr\ program} is a finite set of rules.  
A {\em (generalized) simpagation rule} is of the form
\[r: H_1 \backslash H_2 \Leftrightarrow C | B\]
where $r:$ is an optional {\em name} (a unique identifier) of a rule.
In the rule {\em head} (left-hand side), $H_1$ and $H_2$ are conjunctions of user-defined constraints,
the optional {\em guard} $C |$ is a conjunction of built-in constraints,
and the {\em body} (right-hand side) $B$ is a goal.
A {\em goal} is a conjunction of built-in and user-defined constraints.
A {\em state} is a goal.
Conjunctions are understood as {\em multisets} of their conjuncts.

In the rule, $H_1$ are called the {\em kept constraints}, while $H_2$ are called the {\em removed constraints}.
At least one of $H_1$ and $H_2$ must be non-empty. 
If $H_1$ is empty, the rule corresponds to a {\em simplification rule}, also written
\[s: H_2 \Leftrightarrow C | B.\]
If $H_2$ is empty, the rule corresponds to a {\em propagation rule}, also written
\[p: H_1 \Rightarrow C | B.\]
} 
\end{definition}
In this work, we restrict given CHR programs to rules without built-in constraints in the body except $\true$ and $\false$.
This restriction is necessary as long as built-in constraint solvers do not support the removal of built-in constraints.

\subsection{Abstract Operational Semantics of CHR}\label{sec:chr:semantics}

Computations in CHR are sequences of rule applications. The operational semantics of CHR is given by the state transition system. 
It relies on a structural equivalence between states that abstracts away from technical details in a transition\cite{raiser_betz_fru_equivalence_revisited_chr09,betz2014unified}.

{\em State equivalence} treats built-in constraints semantically and user-defined constraints syntactically.
Basically, two states are equivalent if their built-in constraints are logically equivalent (imply each other) and 
their user-defined constraints form syntactically equivalent multisets.
For example, 
$$X{=<}Y \land Y{=<}X \land c(X,Y) \ \equiv \ X{=}Y \land c(X,X) \not\equiv X{=}Y \land c(X,X) \land c(X,X).$$

For a state $S$, the notation $S_{bi}$ denotes the built-in constraints of $S$
and $S_{ud}$ denotes the user-defined constraints of $S$.
\begin{definition}[State Equivalence]
{\rm
Two states $S_1 = (S_{1bi} \land S_{1ud})$ and $S_2 = (S_{2bi} \land S_{2ud})$ are 
{\em equivalent}, 
written $S_1 \equiv S_2$, if and only if
$$	
\models 
        \forall (S_{1bi} \rightarrow \exists \bar y ((S_{1ud} = S_{2ud}) \land S_{2bi}))
	\land 
         \forall (S_{2bi} \rightarrow \exists \bar x ((S_{1ud} = S_{2ud}) \land S_{1bi}))
$$
\noindent with $\bar x$ those variables that only occur in $S_1$ and $\bar y$ those variables that only occur in $S_2$.
} 
\end{definition}

Using this state equivalence, the abstract CHR semantics is defined by a single transition
(computation step). It defines the application of a rule. 
Note that CHR is a committed-choice language, i.e. there is no backtracking in the rule applications. 
\begin{definition}[Transition]
{\rm
Let the rule $(r : H_1 \backslash H_2 \Leftrightarrow C | B)$ be a variant\footnote{A variant (renaming) of an expression is obtained by uniformly replacing its variables by fresh variables.}
of a rule from a given program $\cal{P}$.
The {\em transition (computation step)}
$S \mapsto_r T$ is defined as follows, where $S$ is called {\em source state} and $T$ is called {\em target state}:
\begin{center}
$\underline{S \equiv (H_1 \land H_2 \land C \land G) \ \ \ (r : H_1 \backslash H_2 \Leftrightarrow C | B) \in {\mathcal{P}}  \ \ \ (H_1 \land C \land B \land G) \equiv T}$\\
$S \mapsto_r T$
\end{center}
The goal $G$ is called {\em context} of the rule application. It is left unchanged. 

A {\em computation (derivation)} of a goal $S$ in a program $\mathcal{P}$
is a connected sequence
$S_i \mapsto_{r_i} S_{i+1}$ beginning with
the {\em initial state (query)} $S_0$ that is $S$ 
and ending in a {\em final state (answer, result)} or the sequence is {\em non-terminating (diverging)}.
We may drop the reference to 
the rules $r_i$ to simplify the presentation.
The notation $\mapsto^*$ denotes the reflexive and transitive closure of $\mapsto$.
} 
\end{definition}
If the source state can be made equivalent to a state that contains the head constraints and the guard built-in constraints of a variant of a rule, then we delete the removed head constraints from the state and add the rule body constraints to it. Any state that is equivalent to this target state is in the transition relation.

The abstract semantics does not account for termination of inconsistent states and propagation rules.
From a state with inconsistent built-in constraints, any transition is possible.
If a state can fire a propagation rule once, it can do so again and again. 
This is called trivial non-termination of propagation rules. 

\medskip
\myparagraph{Minimum Example, contd}
Here is a possible transition from a state $S=(min(0) \land min(2) \land min(1))$ to a state $T=(min(0) \land min(1))$:
\begin{center}
$S \equiv (min(X) \land min(Y) \land X \leq Y \land (X=0 \land Y=2 \land min(1)))$\\ 
$(min(X) \backslash min(Y) \Leftrightarrow X \leq Y | true)$\\ 
$\underline{(min(X) \land X \leq Y \land true \land (X=0 \land Y=2 \land min(1))) \equiv T}$\\
$S \mapsto T$
\end{center}

\section{CHR with Justifications (CHR$^{\mathcal J}$)}

We present a conservative extension of CHR by justifications. If they are not used, programs behave as without them.
Justifications annotate atomic CHR constraints. 
A simple source-to-source transformation extends the rules with justifications. 

\begin{definition}[CHR Constraints and Initial States with Justifications]
{\rm
A {\em justification} $f$  is a unique identifier. 
Given an atomic CHR constraint $G$, 
a {\em CHR constraint with justifications} is of the form
$G^F$, where $F$ is a set of justifications.
An {\em initial state with justifications} is of the form
$\bigwedge_{i=1}^{n} G^{\{f_i\}}_i$
where the $f_i$ are distinct justifications.
} 
\end{definition}

We now define a source-to-source translation from rules to rules with justifications.
Let $\mathit{kill}$ and $\mathit{rem}$ (remove) be to unary {\em reserved} CHR constraint symbols. This means they are only allowed to occur in rules as specified in the following.
\begin{definition}[Translation to Rules with Justifications]
{\rm
Given a generalized simpagation rule
\[r: \bigwedge_{i=1}^{l} K_i \ \backslash \ \bigwedge_{j=1}^{m} R_j \Leftrightarrow C \ | \ \bigwedge_{k=1}^{n} B_k\]
Its translation to a {\em simpagation rule with justifications} is of the form
\[rf: \bigwedge_{i=1}^{l} K^{F_i}_i \ \backslash \ \bigwedge_{j=1}^{m} R^{F_j}_j \Leftrightarrow 
C \ | \ \bigwedge_{j=1}^{m} \mathit{rem}(R^{F_j}_j)^{F} \wedge \bigwedge_{k=1}^{n} B^{F}_k 
\mbox{ where } F = \bigcup_{i=1}^{l} F_i \cup \bigcup_{j=1}^{m} F_j.\]
} 
\end{definition}
The translation ensures that the head and the body of a rule mention exactly the same justifications. 
More precisely, each CHR constraint in the body is annotated with the union of all justifications in the head of the rule, because its creation is caused by the head constraints.
The reserved CHR constraint {\em rem/1} (remember removed) stores the constraints removed by the rule together with their justifications.

\subsection{Operational Equivalence of Rule Applications}

Let $A, B, C \ldots$ be states.
For convenience, we will often consider them as multisets of atomic constraints.
Then the notation $A{-}B$ denotes multiset difference, $A$ without $B$.
By abuse of notation, let $A^{\mathcal J}, B^{\mathcal J}, C^{\mathcal J} \ldots$ be conjunctions and corresponding states whose atomic CHR constraints are annotated with justifications according to the above definition of the rule scheme. Similarly, let $\mathit{rem}(R)^{\mathcal J}$ denote the conjunction $\bigwedge_{j=1}^{m} \mathit{rem}(R^{F_j}_j)^{F}$.

We show that rule applications correspond to each other in standard CHR and in CHR$^{\mathcal J}$.
\begin{lemma}[Equivalence of Program Rules]\label{hugo}
{\rm
There is a computation step $S \mapsto_r T$ with simpagation rule
$$r : H_1 \backslash H_2 \Leftrightarrow C | B$$
if and only if there is a computation step with justifications 
$S^{\mathcal J} \mapsto_{rf} T^{\mathcal J} \wedge \mathit{rem}(H_2)^{\mathcal J}$
with the corresponding simpagation rule with justifications
$$rf : H_1^{\mathcal J} \backslash H_2^{\mathcal J} \Leftrightarrow C | \mathit{rem}(H_2)^{\mathcal J} \land B^{\mathcal J}.$$

\myparagraph{Proof}
We compare the two transitions involving rule {\em r} and {\em rf}, respectively:

\begin{center}
$(r : H_1 \backslash H_2 \Leftrightarrow C | B)$\\
$\underline{S \equiv (H_1 \land H_2 \land C \land G) \ \ \ (H_1 \land C \land B \land G) \equiv T}$\\
$S \mapsto_r T$
\end{center}

\begin{center}
$(rf : H_1^{\mathcal J} \backslash H_2^{\mathcal J} \Leftrightarrow C | \mathit{rem}(H_2)^{\mathcal J} \land B^{\mathcal J})$\\
$\underline{S^{\mathcal J} \equiv (H_1^{\mathcal J} \land H_2^{\mathcal J} \land C \land G^{\mathcal J}) \ \ \  (H_1^{\mathcal J} \land C \land B^{\mathcal J} \land G^{\mathcal J}) \equiv T^{\mathcal J} \land \mathit{rem}(H_2)^{\mathcal J}}$\\
$S^{\mathcal J} \mapsto_{rf} T^{\mathcal J} \land \mathit{rem}(H_2)^{\mathcal J}$
\end{center}

Given the standard transition with rule $r$,
the transition with justifications with rule $\mathit{rf}$ is always possible:
The rule $\mathit{rf}$ by definition does not impose any constraints on its justifications.
The justifications in the rule body are computed as the union of the justifications in the rule head, which is always possible.
Furthermore, the reserved {\em rem} constraints always belong to the context of the transition since by definition 
there is no rule {\em rf} that can match any of them.

Conversely, given the transition with justifications with rule $\mathit{rf}$, by the same arguments,
we can strip away all justifications from it and remove 
$\mathit{rem}(H_2)^{\mathcal J}$ from the rule and the target state to arrive at the standard transition with rule $r$.
\qed}\end{lemma}
Since computations are sequences of connected computation steps, 
this lemma implies that computations in standard CHR program and in CHR$^{\mathcal J}$ correspond to each other.
Thus CHR with justifications is a conservative extension of CHR.

\section{Logical Retraction Using Justifications}

We use justifications to remove a CHR constraint from a computation without the need to recompute from scratch. This means that all its consequences due to rule applications it was involved in are undone.
CHR constraints added by those rules are removed and CHR constraints removed by the rules are re-added.
To specify and implement this behavior, we give a scheme of two rules, one for retraction and one for re-adding of constraints.
The reserved CHR constraint $\mathit{kill}(f)$ undoes all consequences of the constraint with justification $f$.
\begin{definition}[Rules for CHR Logical Retraction]
{\rm
For each $n$-ary CHR constraint symbol $c$ (except the reserved $\mathit{kill}$ and $\mathit{rem}$),
we add a rule to kill constraints and a rule to revive removed constraints of the form:
\[\textrm{kill}: \mathit{kill}(f) \ \backslash \ G^F \Leftrightarrow {f \in F} \ | \ \true\]
\[\textrm{revive}: \mathit{kill}(f) \ \backslash \ rem(G^{F_c})^{F} \Leftrightarrow {f \in F} \ | \ G^{F_c},\]
where $G=c(X_1,\ldots,X_n)$, where $X_1,\ldots,X_n$ are different variables.
} 
\end{definition}
Note that a constraint may be revived and subsequently killed. This is the case when both $F_c$ and $F$ contain the justification $f$.

\subsection{Confluence of Logical Retraction}

Confluence of a program guarantees that any computation starting from a given initial state can always reach equivalent states, no matter which of the applicable rules are applied. There is a decidable, sufficient and necessary syntactic condition to check confluence of terminating programs and to detect rule pairs that lead to non-confluence when applied.

\begin{definition}[Confluence]
{\rm
  If $A \mapsto^* B$ and $A \mapsto^* C$ then there exist states $D_1$ and $D_2$ such that 
$B \mapsto^* D_1$ and $C \mapsto^* D_2$ where $D_1 \equiv D_2$.
} 
\end{definition}

\begin{theorem}\cite{abd_sem_conf_prop_rules_cp97,abd_fru_meuss_confluence_semantics_csr_constr99} \label{conflcp}
{\rm
A terminating CHR\ program is confluent if and only if all its critical pairs are joinable.                  
} 
\end{theorem}
Decidability comes from the fact that there is only a finite number of critical pairs to consider.

\begin{definition}[Overlap, Critical Pair]
{\rm
Given two (not necessarily different) simpagation rules whose variables have been renamed apart,
$K_1 \backslash R_1 \Leftrightarrow C_1 | B_1$ and
$K_2 \backslash R_2 \Leftrightarrow C_2 | B_2$.
Let $A_1$ and $A_2$ be non-empty conjunctions of constraints taken from $K_1 \land R_1$ and $K_2 \land R_2$, respectively.
An {\em overlap} of the two rules is the state consisting of the rules heads and guards:
  $$((K_1 \land R_1) - A_1) \land K_2 \land R_2 \land A_1{\equal}A_2 \land C_1 \land C_2.$$

The {\em critical pair} are the two states that come from applying the two rules to the overlap,
where $E = (A_1{\equal}A_2 \land C_1 \land C_2)$:
  $$(((K_1 \land K_2 \land R_2) - A_2) \land B_1 \land E <>
     ((K_1 \land R_1 \land K_2) - A_1) \land B_2 \land E).$$
} 
\end{definition}
Note that the two states in the critical pair differ by $R_2 \land B_1$ and $R_1 \land B_2$.

A critical pair is {\em trivially joinable} if its built-in constraints are inconsistent or if 
both $A_1$ and $A_2$ do not contain removed constraints \cite{abd_fru_meuss_confluence_semantics_csr_constr99}.

We are ready to show the confluence of the $\mathit{kill}$ and $\mathit{revive}$ rules with each other and with each rule in any given program. It is not necessary that the given program is confluent. This means for any given program, the order between applying applicable rules from the program and retracting constraints can be freely interchanged. It does not matter for the result, if we kill a constraint first or if we apply a rule to it and kill it and its consequences later. 
\begin{theorem}[Confluence of Logical Retraction]\label{conflkr}
{\rm
Given a CHR program whose rules are translated to rules with justifications together with the $\mathit{kill}$ and $\mathit{revive}$ rules.
We assume there is at most one $\mathit{kill}(f)$ constraint for each justification $f$ in any state.
Then all critical pairs between the $\mathit{kill}$ and $\mathit{revive}$ rules and any rule from the program with justifications are joinable.

\medskip
{\bf Proof.} 
There is only one overlap between the $\mathit{kill}$ and $\mathit{revive}$ rules, 
\[\textrm{kill}: \mathit{kill}(f) \ \backslash \ G^F \Leftrightarrow {f \in F} \ | \ \true\]
\[\textrm{revive}: \mathit{kill}(f) \ \backslash \ rem(G^{F_c})^{F} \Leftrightarrow {f \in F} \ | \ G^{F_c},\]
since $G^F$ cannot have the reserved constraint symbol $\mathit{rem/1}$. The overlap is in the $\mathit{kill}(f)$ constraint. But since it is not removed by any rule, the resulting critical pair is trivially joinable.

By our assumption, the only overlap between two instances of the $\mathit{kill}$ rule must have a single $\mathit{kill}(f)$ constraint. Again, since it is not removed, the resulting critical pair is trivially joinable. The same argument applies to the only overlap between two instances of the $\mathit{revive}$ rule.

Since the head of a simpagation rule with justifications from the given program
$$rf : K^{\mathcal J} \backslash R^{\mathcal J} \Leftrightarrow C | \mathit{rem}(R)^{\mathcal J} \land B^{\mathcal J}$$
cannot contain reserved $\mathit{kill}$ and $\mathit{rem}$ constraints, 
these program rules cannot have an overlap with the $\mathit{revive}$ rule. 

But there are overlaps between program rules, say a rule $\mathit{rf}$, and the $\mathit{kill}$ rule.
They take the general form:
$$\mathit{kill}(f) \land K^{\mathcal J} \land R^{\mathcal J} \land G^F{=}A^F \land {f{\in}F} \land C,$$
where $A^F$ occurs in $K^{\mathcal J} \land R^{\mathcal J}$.
This leads to the critical pair
$$(\mathit{kill}(f) \land ((K^{\mathcal J} \land R^{\mathcal J}) - G^F) \land E  <>
\mathit{kill}(f) \land K^{\mathcal J} \land \mathit{rem}(R)^{\mathcal J} \land B^{\mathcal J} \land E),$$
where $E = (G^F{=}A^F \land {f{\in}F} \land C)$.
In the first state of the critical pair, the $\mathit{kill}$ rule has been applied
and in the second state the rule $\mathit{rf}$.
Note that $A^F$ is atomic since it is equated to $G^F$ in $E$.
Since $G^F$ has been removed in the first state and $G^F{=}A^F$,
rule $\mathit{rf}$ is no longer applicable in that state.

We would like to join these two states. The joinability between a rule {\em rf} and the {\em kill} rule can be visualized by the diagram:
{\small
\[
\xymatrix{
  & \hspace*{-1.5em} \mathit{kill}(f) \land K^{\mathcal J}  \land R^{\mathcal J} \land  E \ar@{|->}[dl]_{kill} \ar@{|->}[dr]^{rf} & \\
      \mathit{kill}(f) \land ((K^{\mathcal J} \land R^{\mathcal J}) - G^F) \land E \ar@{<-|}[r]_/4em/{{revive^*,kill^*}}^/-0em/{*}  &  & \hspace*{-7em}
\mathit{kill}(f) \land K^{\mathcal J} \land \mathit{rem}(R)^{\mathcal J} \land B^{\mathcal J} \land E 
}
\]
}

We now explain this joinability result.
The states of the critical pair differ. 
In the first state we have the constraints $R^{\mathcal J}$ and have $G^F$ removed from $K^{\mathcal J} \land R^{\mathcal J}$,
while in the second state we have the body constraints $\mathit{rem}(R)^{\mathcal J} \land B^{\mathcal J}$ of rule $\mathit{rf}$ instead.
Any constraint in $\mathit{rem}(R)^{\mathcal J} \land B^{\mathcal J}$ must include $f$ as justification by definition, because $f$ occurred in the head constraint $A^F$ and $E$ contains ${f{\in}F}$.

The goal $\mathit{rem}(R)^{\mathcal J}$ contains $\mathit{rem}$ constraints for each removed constraint from $R^{\mathcal J}$.
But then we can use $\mathit{kill}(f)$ with the $\mathit{revive}$ rule to replace all $\mathit{rem}$ constraints by the removed constraints, 
thus adding $R^{\mathcal J}$ back again.
Furthermore, we can use $\mathit{kill}(f)$ with the $\mathit{revive}$ rule to remove each constraint in $B^{\mathcal J}$, as each constraint in $B^{\mathcal J}$ contains the justification $f$. 
So $\mathit{rem}(R)^{\mathcal J} \land B^{\mathcal J}$ has been removed completely and $R^{\mathcal J}$ has been re-added. 

The two states may still differ in the occurrence of $G^F$ (which is $A^F$). 
In the first state, $G^F$ was removed by the {\em kill} rule.
Now if $A^F$ ($G^F$) was in $R^{\mathcal J}$, it has been revived with $R^{\mathcal J}$. 
But then the $\mathit{kill}$ rule is applicable and we can remove $A^F$ again.
In the second state, if $A^F$ was in $R^{\mathcal J}$ it has been removed together with $R^{\mathcal J}$ by application of rule {\em rf}. 
Otherwise, $A^F$ is still contained in $K^{\mathcal J}$. But then the $\mathit{kill}$ rule is applicable to $A^F$ and 
removes it from $K^{\mathcal J}$. Now $A^F$ ($G^F$) does not occur in the second state either.

We thus have arrived at the first state of the critical pair.
Therefore the critical pair is joinable.
\qed} 
\end{theorem}
This means that given a state, if there is a constraint to be retracted, we can either kill it immediately or still apply a rule to it and use the $\mathit{kill}$ and $\mathit{revive}$ rules afterwards to arrive at the same resulting state.

Note that the confluence between the $\mathit{kill}$ and $\mathit{revive}$ rules and any rule from the program is independent of the confluence of the rules in the given program.

\subsection{Correctness of Logical Retraction}

We prove correctness of logical retraction:
the result of a computation with retraction is the same as if the constraint would never have been introduced in the computation.
We show that given a computation starting from an initial state with a {\em kill(f)} constraint that ends in a state where 
the {\em kill} and {\em revive} rules are not applicable, i.e. these rules have been applied to exhaustion,
then there is a corresponding computation without constraints that contain the justification $f$.
\begin{theorem}[Correctness of Logical Retraction]\label{hugo1}
{\rm
Given a computation 
$$A^{\mathcal J} \land G^{\{f\}} \land \mathit{kill}(f) \mapsto^{*} B^{\mathcal J} \land \mathit{rem}(R)^{\mathcal J} \land \mathit{kill}(f) \not\mapsto_{kill,revive},$$
where $f$ does not occur in $A^{\mathcal J}$.
Then there is a computation without $G^{\{f\}}$ and $\mathit{kill}(f)$
$$A^{\mathcal J} \mapsto^{*} B^{\mathcal J} \land \mathit{rem}(R)^{\mathcal J}.$$

\medskip
{\bf Proof.} 
We distinguish between transitions that involve the justification $f$ or do not.
A rule that applies to constraints that do not contain the justification $f$ will produce constraints that do not contain the justification.
A rule application that involves at least one constraint with a justification $f$ will only produce constraints that contain the justification $f$.

We now define a mapping from a computation with $G^{\{f\}}$ to a corresponding computation without $G^{\{f\}}$.
The mapping essentially strips away constraints that contain the justification $f$ except those that are remembered by {\em rem} constraints.
In this way, the exhaustive application of the {\em revive} and {\em kill} rules $\mathit{kill}(f)$ is mimicked.

\medskip
$\mathit{strip}(f,A^{\mathcal J} \wedge B^{\mathcal J}) := \ \mathit{strip}(f,A^{\mathcal J}) \wedge \mathit{strip}(f,B^{\mathcal J})$

$\mathit{strip}(f,\mathit{rem}(G^{F_1})^{F_2}) := \ \mathit{strip}(f,G^{F_1})$ if $f \in F_2$

$\mathit{strip}(f,G^F) := \ \true$ if $G$ is an atomic constraint except {\em rem/1} and $f \in F$

$\mathit{strip}(f,G^F) := \ G^F$ otherwise.

\medskip
We extend the mapping from states to transitions.
We keep the transitions except where the source and target state are equivalent, in that case we replace the transition 
$\mapsto$ by an equivalence $\equiv$. This happens when a rule is applied that involves the justification $f$.
The mapping is defined in such a way that in this case the source and target state are equivalent.
Otherwise a rule that does not involve $f$ has been applied. The mapping ensures in this case that all necessary constraints are in the source and target state, since it keeps all constraints that do not mention the justification $f$.
For a computation step $C^{\mathcal J} \mapsto D^{\mathcal J}$ we define the mapping as:

\medskip
$\mathit{strip}(f,C^{\mathcal J} \mapsto_{rf} D^{\mathcal J}) := \ \mathit{strip}(f,C^{\mathcal J}) \equiv \mathit{strip}(f,D^{\mathcal J})$ 
                                if rule $\mathit{rf}$ involves $f$

$\mathit{strip}(f,C^{\mathcal J} \mapsto_{rf} D^{\mathcal J}) := \ \mathit{strip}(f,C^{\mathcal J}) \mapsto_{rf} \mathit{strip}(f,D^{\mathcal J})$ otherwise.

\medskip
We next have to show is that the mapping results in correct state equivalences and transitions.
If a rule is applied that does not involve justification $f$, then it is easy to see that the mapping {\em strip($f,\ldots$)} leaves states and transitions unchanged. 

Otherwise the transition is the application of a rule {\em rf} from the program, the rule {\em kill} or the rule {\em revive}
where $f$ is contained in the justifications.
Let the context $E^{\mathcal J}$ be an arbitrary goal where $f \in \mathcal J$. Then we have to compute
\[\mathit{strip}(f, \mathit{kill}(f) \land G^F \land {f \in F} \land E^{\mathcal J} \mapsto_\mathit{kill} \mathit{kill}(f) \land E^{\mathcal J})\] 
\[\mathit{strip}(f, \mathit{kill}(f) \land \ rem(G^{F_c})^{F} \land {f \in F} \land E^{\mathcal J} \mapsto_\mathit{revive}  \mathit{kill}(f) \land G^{F_c} \land E^{\mathcal J})\] 
\[\mathit{strip}(f, K^{\mathcal J} \land R^{\mathcal J} \land C \land E^{\mathcal J} \mapsto_\mathit{rf} K^{\mathcal J} \land \mathit{rem}(R)^{\mathcal J} \land B^{\mathcal J} \land C \land E^{\mathcal J})\] 
and to show that equivalent states are produced in each case. The resulting states are
\[\true \land \true \land \true  \land E^{\mathcal J'} \equiv \true  \land E^{\mathcal J'}\] 
\[\true \land G^{F_c} \land \true  \land E^{\mathcal J'} \equiv  \true  \land G^{F_c} \land E^{\mathcal J'} \mbox{ if } f \not\in F_c\] 
\[\true \land \true \land \true  \land E^{\mathcal J'} \equiv  \true  \land \true \land E^{\mathcal J'} \mbox{ if } f \in F_c\] 
\[K^{\mathcal J'} \land R^{\mathcal J'} \land C \land E^{\mathcal J'} \equiv K^{\mathcal J'} \land R^{\mathcal J'} \land C \land E^{\mathcal J'} \mbox{ where } f \not\in \mathcal J',\]
where, given a goal $A$, the expression $A^{\mathcal J'}$ contains all constraints from $A^{\mathcal J}$ that do not contain the justification $f$.

\medskip
In the end state of the given computation we know that the {\em revive} and {\em kill} rules have been applied to exhaustion.
Therefore all $\mathit{rem}(G^{F_1})^{F_2}$ where $F_2$ contains $f$ have been replaced by $G^{F_1}$ by the {\em revive} rule.
Therefore all standard constraints with justification $f$ have been removed by the {\em kill} rule (including those revived), 
just as we do in the mapping {\em strip($f,\ldots$)}. 

Therefore the end states are indeed equivalent except for the remaining {\em kill} constraint.
\qed}\end{theorem}

\section{Implementation}

As a proof-of-concept, we implement CHR with justifications (CHR$^{\mathcal J}$) in SWI-Prolog using its CHR library. 
This prototype source-to-source transformation is available online at \url{http://pmx.informatik.uni-ulm.de/chr/translator/}.
The translated programs can be run in Prolog or online systems like WebCHR.

\myparagraph{Constraints with Justifications}
CHR constraints annotated by a set of justifications are realized by a binary infix operator \verb+##+, 
where the second argument is a list of justifications:

\medskip
\noindent $C^{\{F_1,F_2,\ldots\}}$ is realized as 
{\tt C \verb+##+ [F1,F2,...]}.

\medskip
For convenience, we add rules that add a new justification to a given constraint {\tt C}.
For each constraint symbol {\tt c} with arity {\tt n} there is a rule of the form

\medskip
\noindent {\tt addjust @ c(X1,X2,...Xn) <=> c(X1,X2,...Xn) \verb+##+ [\_F].}

\medskip
\noindent where the arguments of {\tt X1,X2,...Xn} are different variables.

\myparagraph{Rules with Justifications}
A CHR simpagation rule with justifications is realized as follows:
\[rf: \bigwedge_{i=1}^{l} K^{F_i}_i \ \backslash \ \bigwedge_{j=1}^{m} R^{F_j}_j \Leftrightarrow 
C \ | \ \bigwedge_{j=1}^{m} \mathit{rem}(R^{F_j}_j)^{F} \wedge \bigwedge_{k=1}^{n} B^{F}_k
\mbox{ where } 
F = \bigcup_{i=1}^{l} F_i \cup \bigcup_{j=1}^{m} F_j\] 

\begin{alltt}
rf @ K1 \verb+##+ FK1,... \verb+\+ R1 \verb+##+ FR1,... <=> C \verb+|+ 
    union([FK1,...FR1,...],Fs), rem(R1\verb+##+FR1) \verb+##+ Fs,...B1 \verb+##+ Fs,...
\end{alltt}
where the auxiliary predicate {\tt union/2} computes the ordered duplicate-free union of a list of lists\footnote{More precisely, a simplification rule is generated if there are no kept constraints and a propagation rule is generated if there are no removed constraints.}.

\myparagraph{Rules {\em remove} and {\em revive}}
Justifications are realized as {\em flags} that are initially unbound logical variables. This eases the generation of new unique justifications and their use in killing. Concretely, the reserved constraint $\mathit{kill}(f)$ is realized as built-in equality {\tt F=r}, i.e. the justification variable gets bound. 
If $\mathit{kill}(f)$ occurred in the head of a {\em kill} or {\em revive} rule, it is moved to the guard as equality test {\tt F==r}. 
Note that we rename rule {\em kill} to {\tt remove} in the implementation.

\medskip
\noindent $\textrm{revive}: \mathit{kill}(f) \ \backslash \ rem(C^{F_c})^{F} \Leftrightarrow {f \in F} \ | \ C^{F_c}$\\
$\textrm{kill}: \mathit{kill}(f) \ \backslash \ C^F \Leftrightarrow {f \in F} \ | \ \true$

\begin{alltt}
revive @ rem(C\verb+##+FC) \verb+##+ Fs <=> member(F,Fs),F==r | C \verb+##+ FC.
remove @ C \verb+##+ Fs <=> member(F,Fs),F==r | true.
\end{alltt}
{Since rules are tried in program order in the CHR implementation, 
the constraint {\tt C} in the second rule is not a reserved {\tt rem/1} constraint when the rule is applicable.}
The check for set membership in the guards is expressed using the standard nondeterministic Prolog built-in predicate {\tt member/2}.

\myparagraph{Logical Retraction with {\tt killc/1}}
We extend the translation to allow for retraction of derived constraints.
The constraint {\tt killc(C)} logically retracts constraint {\tt C}.
The two rules {\tt killc} and {\tt killr} try to find the constraint {\tt C} - also when it has been removed and is now present in a {\tt rem} constraint.
The associated justifications point to all initial constraints that where involved in producing the constraint {\tt C}. For retracting the constraint, it is sufficient to remove one of its producers. This introduces a choice implemented by the {\tt member} predicate.
\begin{alltt}
killr @  killc(C), rem(C ## FC) ## _Fs <=> member(F,FC),F=r.  
killc @  killc(C), C ## Fs <=> member(F,Fs),F=r.
\end{alltt}
Note that in the first rule, we bind a justification {\tt F} from {\tt FC}, because {\tt FC} contains the justifications of the producers of constraint {\tt C}, while {\tt Fs} also contains those that removed it by a rule application.

\subsection{Examples}

We discuss two classical examples for dynamic algorithms,
maintaining the 
minimum of a changing set of numbers and shortest paths when edges change.

\myparagraph{Dynamic Minimum}
Translating the minimum rule to one with justifications results in:
\begin{alltt}
min(A)##B \verb+\+ min(C)##D <=> A<C | union([B,D],E), rem(min(C)##D)##E.
\end{alltt}
The following shows an example query and the resulting answer in SWI-Prolog:
\begin{alltt}
?- min(1)##[A], min(0)##[B], min(2)##[C].
rem(min(1)##[A])##[A,B], rem(min(2)##[C])##[B,C],
min(0)##[B].
\end{alltt}
The constraint {\tt min(0)} remained. This means that {\tt 0} is the minimum. 
The constraints {\tt min(1)} and {\tt min(2)} have been removed and are now remembered.
Both have been removed by the constraint with justification {\tt B}, i.e. {\tt min(0)}.

We now logically retract with {\tt killc} the constraint {\tt min(1)} at the end of the query. 
The {\tt killr} rule applies and removes {\tt rem(min(1)\#\#[A])\#\#[A,B]}. 
(In the rule body, the justification {\tt A} is bound to {\tt r} - 
to no effect, since there are no other constraints with this justification.)
\begin{alltt}
?- min(1)##[A], min(0)##[B], min(2)##[C], killc(min(1)).
rem(min(2)##[C])##[B,C],
min(0)##[B].
\end{alltt}

What happens if we remove the current minimum {\tt min(0)}?
The constraint {\tt min(0)} is removed by binding justification {\tt B}.
The two {\tt rem} constraints for {\tt min(1)} and {\tt min(2)} involve {\tt B} as well,
so these two constraints are re-introduced and react with each other. 
Note that  {\tt min(2)} is nwo removed by {\tt min(1)} (before it was {\tt min(0)}).
The result is the updated minimum, which is {\tt 1}.
\begin{alltt}
?- min(1)##[A], min(0)##[B], min(2)##[C], killc(min(0)).
rem(min(2)##[C])##[A,C],
min(1)##[B].
\end{alltt}

\myparagraph{Dynamic Shortest Path}
Given a graph with directed arcs {\tt e(X,Y)}, we compute the lengths of the shortest paths between all pairs of reachable nodes:
\begin{alltt}
 % keep shorter of two paths from X to Y
pp @ p(X,Y,L1) \verb+\+ p(X,Y,L2) <=> L1=<L2 | true.
 % edges have a path of unit length
e  @ e(X,Y) ==> p(X,Y,1).
 % extend path in front by an edge
ep @ e(X,Y), p(Y,Z,L) ==> L1=:=L+1 | p(X,Z,L1).
\end{alltt}
The corresponding rules in the translated program are:
\begin{alltt}
pp@p(A,B,C)##D \verb+\+ p(A,B,E)##F <=> C=<E | 
                             union([D,F],G), rem(p(A,B,E)##F)##G.
e @e(A,B)##C ==> true | union([C],D), p(A,B,1)##D.
ep@e(A,B)##C,p(B,D,E)##F ==> G is E+1 | union([C,F],H),p(A,D,G)##H.
\end{alltt}

We now use constraints without justifications in queries. Justifications will be added by the {\tt addjust} rules.
\begin{alltt}
?- e(a,b), e(b,c), e(a,c).
rem(p(a, c, 2)##[A, B])##[A,B,C],
p(a, b, 1)##[A], e(a, b)##[A],
p(b, c, 1)##[B], e(b, c)##[B],
p(a, c, 1)##[C], e(a, c)##[C].
\end{alltt}
We see that a path of length {\tt 2} has been removed by the constraint {\tt e(a,c)\verb+##+[C]}, 
which produced a shorter path of length one.
We next kill this constraint {\tt e(a,c)}.
\begin{alltt}
?- e(a,b), e(b,c), e(a,c), kill(e(a,c)).
p(a, b, 1)##[A], e(a, b)##[A],
p(b, c, 1)##[B], e(b, c)##[B],
p(a, c, 2)##[A,B].
\end{alltt}
Its path {\tt p(a,c,1)} disappears and the removed remembered path {\tt p(a,c,2)} is re-added.
We can see that the justifications of a path contains are those from the edges in that path.
The same happens if we logically retract {\tt p(a,c,1)} instead of {\tt e(a,c)}.

What happens if we remove {\tt p(a,c,2)} from the initial query? 
The {\tt killr} rule applies. Since the path has two justifications, 
there are two computations generated by the {\tt member} predicate. In the first one, the constraint 
{\tt e(a,b)} disappeared, in the second answer, it is {\tt e(b,c)}. 
In both cases, the path cannot be computed anymore, i.e. it has been logically retracted.
\begin{alltt}
?- e(a,b), e(b,c), e(a,c), kill(p(a,c,2)).
p(b, c, 1)##[B], e(b, c)##[B],
p(a, c, 1)##[C], e(a, c)##[C]
;   
p(a, b, 1)##[A], e(a, b)##[A],
p(a, c, 1)##[C], e(a, c)##[C].
\end{alltt}

\section{Related Work}

The idea of introducing justifications into CHR is not new. The thorough work of Armin Wolf
on Adaptive CHR \cite{wolf2000incremental} was the first to do so. Different to our work, this technically involved approach requires to store detailed information about the rule instances that have been applied in a derivation in order to undo them.
Adaptive CHR had a low-level implementation in Java \cite{wolf2001adaptive}, 
while we give an implementation in CHR itself by a straightforward source-to-source transformation 
that we prove confluent and correct. 
Moreover we prove confluence of the rule scheme for logical retraction with the rules of the given program. 
The application of adaptive CHR considered dynamic constraint satisfaction problems (DCSP) only, 
in particular for the implementation of search strategies \cite{wolf2005intelligent},  
while we apply our approach to arbitrary algorithms in order to make them fully dynamic.

The issue of search strategies was further investigated by
Leslie De Koninck et. al. \cite{de2008flexible}. 
They introduce a flexible search framework in CHR$^\lor$ (CHR with disjunction)
extended with rule and search branch priorities. 
In their work, justifications are introduced into the semantics of CHR$^\lor$ to enable 
dependency-directed backtracking in the form of conflict-directed backjumping.
Our work does not need a new semantics for CHR, nor its extension with disjunction, 
it rather relies on a source-to-source transformation within the standard semantics.

Our work does not have a particular application of justifications in mind, but rather provides the basis for any type of application that requires dynamic algorithms. There are, however, CHR applications that use justifications.

The work of Jeremy Wazny et. al. \cite{stuckey2003interactive} introduced informally a particular kind of justifications into CHR for the specific application of type debugging and reasoning in Haskell. Justifications correspond to program locations in the given Haskell program. Unlike other work, the constraints in the body of CHR rules have given justifications to which justifications from the rule applications are added at runtime. 

The more recent work of Gregory Duck \cite{duck2012smchr} introduces SMCHR, 
a tight integration of CHR with a Boolean Satisfiability (SAT) solver for quantifier-free formulae including disjunction and negation as logical connectives. It is mentioned  that for clause generation, SMCHR supports justifications for constraints that include syntactic equality constraints between variables. A dynamic unification algorithm using justifications has been considered in \cite{Wolf1998}.

\section{Conclusions}

In this paper, the basic framework for CHR with justifications (CHR$^{\mathcal J}$) has been established and formally analyzed.
We defined a straightforward source-to-source program transformation that introduces justifications into CHR as a conservative extension. Justifications enable logical retraction of constraints. If a constraint is retracted, the computation continues as if the constraint never was introduced. We proved confluence and correctness of the two-rule scheme that encodes the logical retraction. We presented a prototype implementation that is available online together with two classical examples.

Future work could proceed along three equally important lines: investigate implementation, dynamic algorithms and application domains of CHR with justifications.
First, we would like to research how logical as well as classical algorithms implemented in CHR behave when they become dynamic.
While our approach does not require confluence of the given program, the arbitrary re-introduction of removed constraints may lead to unwanted orders of rule applications in non-confluent programs.
Second, we would like to improve the implementation, optimize and benchmark it
\cite{frueh17inap}.
Currently, the entire history of removed constraints is stored. It could suffice to remember only a partial history if only certain constraints can be retracted or if partial recomputation proves to be more efficient for some constraints. A lower-level implementation could benefit from the propagation history that comes with the application of propagation rules in most CHR implementations.
Third, we would like to extend the rule scheme to support typical application domains of justifications: 
explanation of derived constraints by justifications (for debugging),
detection and repair of inconsistencies (for error diagnosis), and
implementing nonmonotonic logical behaviors (e.g. default logic, abduction, defeasible reasoning).
Logical retraction of constraints can also lead to reversal of computations, and as a first step the related literature on reversibility should be explored.

{\bf Acknowledgements.} 
We thank Daniel Gall for implementing the online transformation tool for CHR$^{\mathcal J}$.
We also thank the anonymous reviewers for their helpful suggestions on how to improve the paper.

\bibliographystyle{alpha} 
\bibliography{chrjust,devils,biblio} 

\label{lastpage}

\end{document}